# EXPERIMENTING ACTIVE AND SEQUENTIAL LEARNING IN A MEDIEVAL MUSIC MANUSCRIPT

*Sachin Sharma, Federico Simonetta, Michele Flammini*

GSSI – Gran Sasso Science Institute

L'Aquila, Italy

## ABSTRACT

Optical Music Recognition (OMR) is a cornerstone of music digitization initiatives in cultural heritage, yet it remains limited by the scarcity of annotated data and the complexity of historical manuscripts. In this paper, we present a preliminary study of Active Learning (AL) and Sequential Learning (SL) tailored for object detection and layout recognition in an old medieval music manuscript. Leveraging YOLOv8, our system selects samples with the highest uncertainty (lowest prediction confidence) for iterative labeling and retraining. Our approach starts with a single annotated image and successfully boosts performance while minimizing manual labeling. Experimental results indicate that comparable accuracy to fully supervised training can be achieved with significantly fewer labeled examples. We test the methodology as a preliminary investigation on a novel dataset offered to the community by the Anonymous project, which studies laude, a poetical-musical genre spread across Italy during the 12th-16th Century. We show that in the manuscript at-hand, uncertainty-based AL is not effective and advocates for more usable methods in data-scarcity scenarios.

***Index Terms***— Optical Music Recognition, Object Detection, Active Learning, YOLOv8, Cultural Heritage

## 1 Introduction

The digitization of historical musical manuscripts through Optical Music Recognition (OMR) is vital for creating searchable archives, digital editions, and advancing computational musicology. OMR automates the extraction of musical symbols from scanned scores. The ultimate utility of such applications often depends on how well the extracted information aligns with human musical understanding, an area where evaluating transcriptions becomes crucial [1].

However, OMR for historical manuscripts faces persistent challenges. The annotation process is exceptionally labor-intensive, requiring domain expertise for dense, often overlapping symbols. Furthermore, there's a scarcity of robust, user-friendly annotation and training tools tailored to computational musicology [2], often compelling researchers to develop custom solutions, although emerging frameworks aim to ease dataset compilation [3]. Finally, historical documents introduce complexities like physical degradation and diverse notational practices, differing significantly from modern scores, as evidenced in projects tackling large historical archives [4]. These issues collectively hinder the development of high- performing, generalizable OMR systems.

This paper attempts to address these challenges by integrating an uncertainty-based Active Learning (AL) strategy into an object detection pipeline for OMR, focusing on Italian medieval lauda manuscripts from the *Anonymous* project. We use YOLOv8, iteratively improving its performance from a small labeled set by selecting uncertain samples for annotation, aiming to reduce labeling costs and enhance robustness. However, when compared with more trivial strategies such as Sequential Learning (SL), the manuscript at hand reveals the inadequacy of the AL strategy. The main contributions are *a)* a preliminary study of AL and SL in OMR, and *b)* a novel annotated medieval musical manuscript dataset, distinct in lay- out and degradation from typically available resources which promises to challenge SOTA AL methods.[1]

## 2 Related Work

Optical Music Recognition (OMR) has evolved from traditional rule-based systems to modern deep learning (DL) approaches, which are crucial for tackling the complexities of music notation, especially in historical scores [5]. Techniques such as Convolutional Neural Networks (CNNs) in various architectures like YOLO are now standard for tasks like symbol detection and layout analysis [6, 7, 8]. These OMR advancements build upon foundational progress in areas like deep feature learning, whose principles are explored across many scientific domains [9]. Despite significant progress, accurately handling notational variability and reconstructing symbol relationships in historical documents remain key challenges [10, 8].

A major bottleneck for applying DL to OMR, particularly for historical manuscripts, is the scarcity of large-scale annotated datasets; their creation is labor-intensive and costly [5]. Ac-tive Learning (AL) offers a promising strategy to mitigate this by iteratively selecting the most informative unlabeled samples for annotation, thereby optimizing the training process and reducing manual effort [11, 12]. Various AL query strategies—focusing on uncertainty, representativeness, or hybrid approaches—have been adapted for deep learning, including methods for object-level selection within images [11, 13, 14]. However, despite successes in other domains, the application of AL for object detection in the OMR context, especially for complex historical manuscripts, remains largely underexplored [12]. This presents a significant opportunity for advancing semi-automated digitization pipelines for musical heritage.

---

[1]The code used in our experiments is available at: https://github.com/LaudareProject/lauda_dataset. The annotated image dataset is **not yet public** and will be released after the LAUDARE project concludes. All code is shared under the *CC BY 4.0 License*.

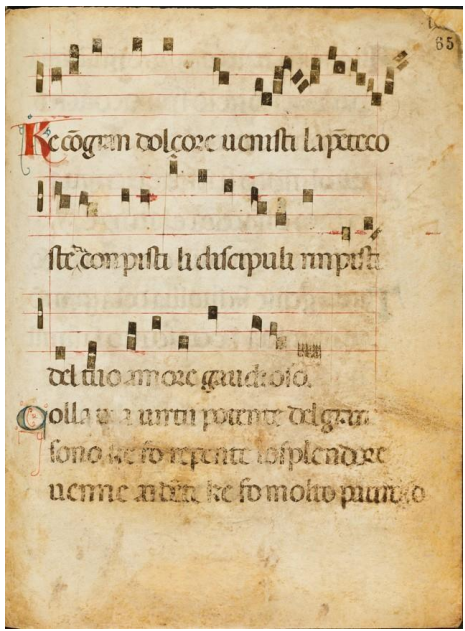

**Fig. 1**: Example of page from the I-Ct 91 manuscript

## 3 Methodology

### 3.1 Dataset

Our experiments are based on the I-Ct 91 manuscript (commonly known as "Cortonese"), a medieval source housed in the Cortona Library and annotated as part of the Anonymous project.

To understand the specificities of this manuscript, it is pertinent to define Lauda. Lauda is a poetical-musical genre prevalent throughout the Italian peninsula between the 12th and 16th centuries. It was a popular genre, sung by "confraternite"—lay religious associations, who performed religious texts in non-liturgical settings. Primarily an oral tradition due to its popular nature, Lauda was occasionally transcribed, as exemplified by the manuscript I-Ct 91. When analyzing this repertoire, it is important to consider that scribes and readers may have possessed limited literacy in music and text, with oral transmission remaining the predominant mode of dissemination.

Given these circumstances, the manuscript exhibits several key characteristics. It incorporates both music and text, with lines of text positioned beneath tetragrams at the commencement of each new lauda; following the initial stanza, the manuscript proceeds exclusively with text. Distinct from typical Gregorian chants, there is no explicit alignment between the musical notation and the corresponding text. The notation employed is non-mensural; consequently, the musical signs, termed "neumes," do not convey durational values. Furthermore, vertical lines, resembling bar delimiters in Common Western Music Notation (CWMN), are occasionally present. These may signify potential word or verse endings; however, their precise interpretation is a subject of debate among experts. Experts have identified multiple different writing styles at both the graphical and notational levels, indicating multiple scribes — a common challenge that complicates computational style analysis in historical music [15].

The dataset annotation was conducted by a collaborative team from the Anonymous project, comprising two musicologists, one philologist, and one researcher specializing in computer science and computational musicology. For the purpose of this work, the annotation was limited to rectangular bounding boxes identifying text regions, music-text regions, text lines, and musical objects. Three distinct software tools were employed for precise annotation.

Initially, Transkribus[2], utilizing both its online and offline graphical user interfaces, was used to annotate text lines, tetragrams, text regions and music-text regions, which produced PageXML files. Subsequently, Neon[3], a specialized software for non-mensural music notation [16], was employed to annotate tetragrams and all relevant musical objects, creating MEI files. Finally, due to imprecisions in Neon's bounding-box encoding, the computer scientist annotated all musical symbols in Inkscape[4] by creating rectangular annotations and saving them as SVG files.

These three annotation sources were then programmatically merged. This process involved matching bounding boxes to align SVG rectangles with the most similar MEI bounding box and to ensure correspondence between tetragrams in PageXML and MEI files. The result of this procedure was a COCO JSON file representing all identified visible objects. Overall, the dataset comprises 7,015 bounding box annotations, with an average of approximately 20.63 annotations per image across 340 images. The dataset comprises 9 categories of musical and textual symbols – see Table 1. We resized all images to 640 × 640 pixels and used YOLO-format for the annotations.

The category designated as "discard" was introduced because not all visible objects within the manuscript images possess semantic meaning relevant to the transcription process. Objects were assigned to this category for several reasons. These include instances where the scribe intentionally deleted signs or markings. Additionally, some visible elements were deliberately not transcribed by the musicologists and philologists due to their illegibility, making accurate interpretation impossible. Finally, certain signs were deemed incorrect by the domain experts, likely resulting from the scribes' limited literacy. Consequently, all such cases were consolidated into this overarching category, indicating that the associated visible objects should not be considered during the subsequent transcription stages.

---

[2][https://www.transkribus.org/](https://www.transkribus.org/)
[3][https://ddmal.music.mcgill.ca/Neon/](https://ddmal.music.mcgill.ca/Neon/)
[4][https://inkscape.org/](https://inkscape.org/)

| Class | Count |
|---|---|
| neume | 2,745 |
| line | 2,522 |
| discard | 530 |
| staff | 301 |
| clef | 261 |
| musicDelimiter | 183 |
| text | 189 |
| custos | 172 |
| musicText | 112 |

**Table 1**: Distribution of annotations per class

## 3.2 Sequential and Active Learning Strategy

In a typical annotation procedure, philologists and musicologists annotate the manuscript starting from the beginning till the end. During this process, they may use machine-learning models to learn from the already transcribed pages and predict the transcription of the next pages, in the hope that correcting the prediction requires less effort than transcribing from scratch. This may be performed iteratively, for instance annotating 10 pages, then training the models and inferencing the next 10 pages, correcting them, re-training the models and so on. The choice of the images that should be annotated at each iteration often relies on the page order, thus proceeding from the first to the last. This procedure is termed "Sequential Learning" (SL).

More advanced strategies have been proposed, based on choosing each iteration image set by optimizing the model's capacity to learn from the input feature space. In practice, the process begins with the initial annotation of a few images, followed by the selection of the subsequent images based on the model's prediction uncertainty. It is anticipated that greater performance improvements can be achieved by incorporating samples into the training set for which the model exhibits higher uncertainty. This strategy is named "Active Learning" (AL)– see Fig. 2.

In the AL procedure, in each iteration, the following steps are executed:

1. The target model is trained on the currently labeled dataset.
2. Inference is performed on the remaining unlabeled images using the trained model.
3. The 15 images exhibiting the *lowest maximum class confidence* are selected, as this value indicates the highest model uncertainty.

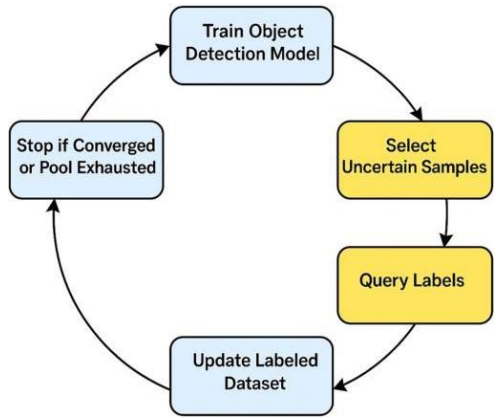

**Fig. 2**: Uncertainty-based active learning cycle

4. These selected images are then manually annotated.
5. The newly annotated images are added to the labeled dataset, and the process continues to the subsequent round.

This iterative approach allows us to gradually expand the labeled dataset with the most informative samples. The method is lightweight, scalable, and particularly suitable in domains like Optical Music Recognition (OMR), where expert annotation is time-consuming and scarce.

## 3.3 YOLOv8 for Object Detection

YOLOv8 (You Only Look Once, version 8) is a single-stage object detection model that performs real-time predictions by processing the entire image in a single forward pass. We utilize YOLOv8n, the lightweight variant of YOLOv8 from Ultralytics[5], an implementation for object detection, fine-tuned with available labeled data. YOLOv8n was chosen over heavier alternatives (e.g. YOLOv8s / m / l or Faster R-CNN) due to its efficiency (3 million parameters, 8.2 GFLOPs) and suitability for low-resource training setups, while still offering strong detection capabilities.

YOLOv8 incorporates notable advancements, including anchor-free detection, which directly predicts object centers and dimensions. This method simplifies training and is advantageous for detecting irregularly shaped objects like music symbols. A decoupled head separates classification and regression, reducing task interference and improving both localization and classification, vital for distinguishing similar classes. The enhanced backbone, with a Cross Stage Partial (CSP)- inspired design and C2f (a faster CSP bottleneck) layers, promotes better gradient flow and feature reuse, crucial for fine details in historical documents. Simplified post-processing, featuring improved Non-Maximum Suppression (NMS) to filter redundant detections, yields cleaner predictions.

YOLOv8's optimization relies on a composite loss function. This includes the Complete Intersection over Union (CIoU) box regression loss for penalizing misalignment, an object- ness loss for confidence in object presence, a classification loss for incorrect labels, and the Distribution Focal Loss (DFL) for modeling box boundary precision via distributional learning.

---
[5]https://github.com/ultralytics/ultralytics

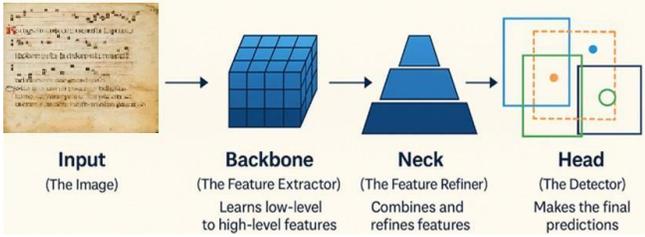

**Fig. 3**: YOLOv8 architecture for object detection

### 3.4 Experimental Setup

In order to fairly evaluate the model performance across different train set size, we selected 20% of the dataset as test set and keep it constant across all the experiments. Specifically, we use a VGG16 network pre-trained on ImageNet-1k to ex- tract features and apply single-linkage clustering based on co- sine similarity. We stop the agglomerative clustering when the number of clusters is equal to 68 (20% of the dataset size). We then select one sample from each cluster to create the test set, leaving 272 images in the train set. This stratified sampling procedure allows to create a test set that optimally represent the differences among the pages and the scribes of the manuscript. We conduct 20 rounds of AL and SL, starting with one labeled image in round 0 and progressively adding 15 new annotated images per round, reaching a total of 272 labeled images in round 19.

The YOLOv8n (nano) model, provided by Ultralytics, is initialized with pretrained weights from the COCO 2017 dataset, which contains 80 object classes. These weights serve as a general-purpose feature extractor and are fine-tuned on our domain-specific dataset for 150 epochs per active learning round. Fine-tuning is applied to the entire architecture— including the backbone, neck, and detection head—with selective freezing (e.g., Distribution Focal Loss (DFL) convolution layers) to retain stable, low-level features. Training is conducted using Automatic Mixed Precision (AMP) and the AdamW optimizer (learning rate $\approx 7.7 \times 10^{-4}$) to accelerate convergence and improve performance.

Prior of training, images are converted to grayscale, to reduce computational overhead and because color information is non-informative for symbol detection in historical manuscripts. We ensure the network focuses solely on spatial and structural features. During inference, the model processes `640 × 640` grayscale manuscript images and outputs detections in YOLO format: `class ID, x center, y center, width, height`, with all coordinates normalized between 0 and 1. We assess object detection performance using standard metrics, including Precision, Recall, and the F1-score. We also report mean Average Precision (mAP) at a 0.5 IoU threshold (mAP@50) and averaged across IoU thresholds from 0.5 to 0.95 (mAP@50-95).

Finally, we set a baseline by fine-tuning the same pre-trained YOLOv8n model on the full 272 train image set. This baseline corresponds to what could be acheived by an extensive annotation of each page from scratch.

## 4 Results and Discussion

The performance of the Active Learning (AL) model across 20 rounds is summarized in Table 2. The AL process began with a single labeled image in Round 0, where the model exhibited negligible performance—mAP@50 was just 0.2%, and both recall and F1-score were effectively 0.0%. By Round 1, precision reached 100%, but recall remained at 0.0%, indicating the model made a few correct predictions but failed to identify most relevant instances.

As more labeled data were added—15 low-confidence sam- ples per round—a consistent improvement was observed. By Round 5 (with 76 images), mAP@50 had increased to 45.0%, though recall was still at 0.0%. The turning point in recall came around Round 6, where it jumped to 24.6%, with a corresponding rise in F1-score to 28.4%. From that point, the model performance improved significantly, achieving an mAP@50 of 76.3%, mAP@50:95 of 52.1%, precision of 70.7%, and recall of 89.6% by Round 16 (241 images), when the labeled pool was nearly exhausted. The final round (Round 19, 272 im- ages) saw the highest AL performance: mAP@50 of 77.3%, mAP@50:95 of 52.4%, precision of 80.7%, recall of 85.8%, and F1-score of 83.2%.

In contrast, the Sequential Learning (SL) strategy showed stronger early performance. At Round 3, SL achieved an mAP@50 of 40.9% and mAP@50:95 of 25.7%, compared to AL's 35.1% and 17.4%, respectively—a +5.8% and +8.3% gain. The difference was especially notable in Round 5, where SL outperformed AL in recall and F1-score, achieving 67.9% and 69.1%, respectively, while AL remained at 0.0% for both. In terms of precision, AL actually performed better with 100.0%, compared to SL's 70.4%.

However, as rounds progressed, the advantage of SL diminished. By Round 19, the difference between SL and AL had nearly vanished. SL achieved an mAP@50 of 77.4%, mAP@50:95 of 53.3%, precision of 85.7%, recall of 88.1%, and F1-score of 86.9%—only slightly higher than AL in most metrics, and actually lower in recall and F1-score in several intermediate rounds.

A fully supervised baseline YOLOv8n model fine-tuned on the complete training set (272 images) achieved mAP@50 of 74% and mAP@50:95 of 47%, lower than the final AL and SL models—highlighting the effectiveness of both sample- efficient approaches. We note a limitation in our baseline com- parison: the fully supervised model is trained once for 150 epochs, while the iterative AL and SL models undergo a total of 3000 epochs of training (20 rounds × 150 epochs). This greater number of optimization steps may contribute to their superior final performance. Future work should explore base- lines trained for a comparable number of total updates.

The initial strong performance of SL could be attributed to

| Round | #Images | Uncertainty-based Active Learning | | | | | Sequential Learning | | | | |
|---|---|---|---|---|---|---|---|---|---|---|---|
| | | mAP@50 | mAP@50:95 | Precision | Recall | F1-score | mAP@50 | mAP@50:95 | Precision | Recall | F1-score |
| 0 | 1 | **0.2%** | **0.1%** | 0.0% | 0.0% | NaN | **0.2%** | **0.1%** | 0.0% | 0.0% | NaN |
| 1 | 16 | **5.9%** | **1.1%** | 100.0% | 0.0% | 0.0% | 3.0% | 0.6% | 100.0% | 0.0% | 0.0% |
| 2 | 31 | 26.7% | 11.9% | 100.0% | 0.0% | 0.0% | **31.4%** | **15.4%** | 100.0% | 0.0% | 0.0% |
| 3 | 46 | 35.1% | 17.4% | 0.0% | 0.0% | NaN | **40.9%** | **25.7%** | 100.0% | 0.0% | 0.0% |
| 4 | 61 | 37.9% | 20.8% | 0.0% | 0.0% | NaN | **54.6%** | **35.9%** | 58.6% | 33.8% | 42.8% |
| 5 | 76 | 45.0% | 24.6% | 100.0% | 0.0% | 0.0% | **62.2%** | **41.3%** | 70.4% | 67.9% | 69.1% |
| 6 | 91 | 55.7% | 35.1% | 33.6% | 24.6% | 28.4% | **65.6%** | **43.9%** | 80.9% | 75.9% | 78.4% |
| 7 | 106 | **67.5%** | 44.6% | 49.2% | 73.1% | 58.8% | 67.2% | **45.9%** | 75.9% | 79.8% | 77.8% |
| 8 | 121 | **69.6%** | **47.1%** | 80.4% | 73.5% | **76.8%** | 68.3% | 46.9% | 66.1% | 87.3% | 75.2% |
| 9 | 136 | **71.5%** | **48.0%** | 83.1% | 73.5% | **78.0%** | 69.8% | 47.5% | 90.8% | 67.2% | 77.2% |
| 10 | 151 | **72.2%** | **48.9%** | 85.3% | 77.9% | 81.5% | 71.1% | 48.6% | 81.8% | 82.8% | 82.3% |
| 11 | 166 | **73.9%** | **50.6%** | 77.3% | **82.1%** | 79.6% | 70.8% | 48.5% | 90.0% | 80.4% | 84.9% |
| 12 | 181 | **74.0%** | **50.9%** | 76.5% | 82.8% | 79.6% | 72.1% | 49.0% | 80.5% | 88.8% | 84.5% |
| 13 | 196 | **75.7%** | **51.5%** | 76.8% | 80.6% | 78.7% | 73.2% | 49.6% | 90.9% | 85.8% | 88.3% |
| 14 | 211 | 75.5% | 52.0% | 76.5% | 82.8% | 79.5% | **75.8%** | **52.2%** | 83.3% | 87.3% | 85.3% |
| 15 | 226 | **76.7%** | 52.3% | 86.6% | **87.3%** | 87.0% | 75.7% | **52.7%** | 87.3% | 87.3% | 87.3% |
| 16 | 241 | 76.3% | 52.1% | 70.7% | **89.6%** | 79.0% | **76.5%** | **52.9%** | 85.8% | 85.6% | 85.7% |
| 17 | 256 | 76.1% | **52.8%** | 84.2% | **88.1%** | 86.1% | **77.5%** | **52.8%** | 90.1% | 87.9% | **88.9%** |
| 18 | 271 | 76.6% | 52.5% | 84.6% | **87.3%** | 85.9% | **77.2%** | **53.0%** | 89.7% | 85.8% | 87.7% |
| 19 | 272 | 77.3% | 52.4% | **80.7%** | 85.8% | 83.2% | **77.4%** | **53.3%** | 85.7% | **88.1%** | 86.9% |

**Table 2**: Numerical results of the Uncertainty-based Active Learning (AL) and Sequential Learning (SL) strategies. Bold values highlight the best performance at each iteration for each metric.

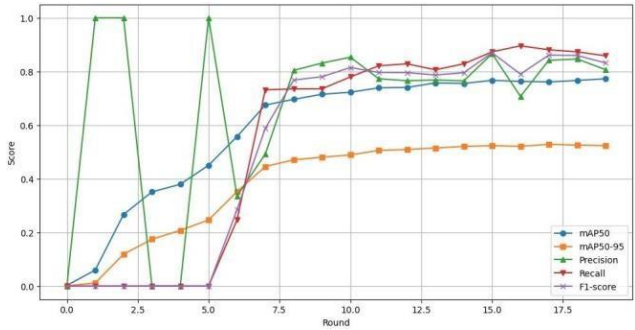

(a) Model Performance Over Active Learning Rounds.

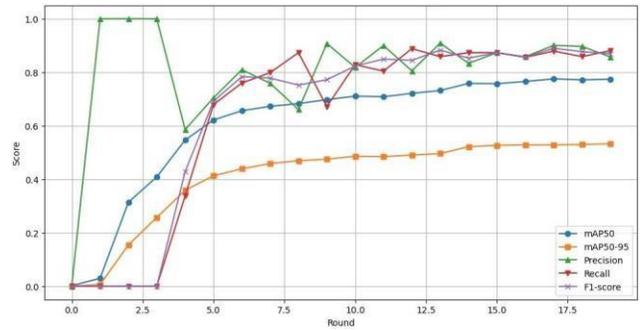

(b) Model Performance Over Sequential Learning Rounds.

**Fig. 4**: Comparison of Active Learning (left) and Sequential Learning (right). SL shows stronger initial performance, particularly in recall and F1-score, while AL gradually closes the performance gap in later rounds.

the multiple scribes that likely wrote the I-Ct 91 manuscript, leading to different graphical a notational properties. Specifically, the early page's style can be tracked to the most preva- lent scribe in the manuscript, which is also the most prevalent in the test set – despite the stratified sampling. In contrast, AL, using confidence-based uncertainty, sampled across varied regions, slowing early learning. This imbalance caused initial performance gaps. However, AL's focus on uncertain, diverse samples allowed it to generalize better and close the gap over time.

This highlights a key vulnerability of simple uncertainty sampling: it can be misled by dataset – specific properties like scribal prevalence and risks developing model bias if it be- comes overconfident in early, incorrect predictions – a pitfall less likely to affect SL. Nevertheless, the dynamic shifts in Table 2, where AL's performance begins to match SL's F1-score around rounds 8-10, suggest a turning point. Once the model has seen a sufficient variety of initial data, AL's targeted selec- tion becomes more beneficial. This indicates that while naive uncertainty sampling is suboptimal here, more advanced AL strategies could offer a more robust solution for such heterogeneous historical documents.

These findings bear direct relevance to the objectives of the Anonymous Project, which endeavors to compile an extensive database of medieval lauda. This research illustrates the proficient application of contemporary AI techniques to cultural heritage studies, thereby facilitating efficient and scalable musicological analysis of intricate historical sources.

## 5 Conclusion and Future Work

We proposed an SL/AL framework for object detection in medieval music manuscripts, specifically applied to the I-Ct 91 ("Cortonese") dataset using the YOLOv8 detector. To the best of our knowledge, this is the first application of AL for object detection in OMR within the cultural heritage domain.

Future works include extending this framework to multi-page layout analysis and exploring domain adaptation across diverse manuscript sources to support semi-automated, large-scale digitization workflows under the Anonymous project.
This work advances AI-assisted cultural heritage preservation and analysis, combining machine learning with the specific needs of musicological research.

## Acknowledgement

This work has been funded by the European Union (Horizon Programme for Research and Innovation 2021-2027, ERC Advanced Grant "The Italian Lauda: Disseminating Poetry and Concepts Through Melody (12th-16th century)", acronym LAUDARE, project no. 101054750). The views and opinions expressed are, however, only those of the author and do not necessarily reflect those of the European Union or the European Research Council. Neither the European Union nor the awarding authority can be held responsible for such matters.

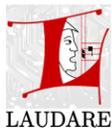
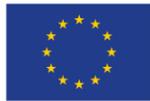
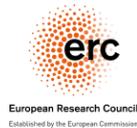